\documentclass[11pt,letterpaper,logo]{thuair}

\usepackage{hyperref}[citecolor=magenta]
\usepackage{fdsymbol}
\usepackage{kantlipsum, lipsum}
\usepackage{dsfont}
\usepackage{subfigure}
\usepackage{bbm}
\usepackage{wrapfig}
\usepackage[most,skins,theorems]{tcolorbox}
\usepackage{authblk}
\usepackage[utf8]{inputenc}
\usepackage[T1]{fontenc}
\usepackage{hyperref}
\usepackage{url}
\usepackage{booktabs}
\usepackage{amsfonts}
\usepackage{nicefrac}
\usepackage{microtype}
\usepackage{lipsum}
\usepackage{fancyhdr}
\usepackage{graphicx}
\usepackage{xspace}
\usepackage{color}
\usepackage{xcolor}
\usepackage{pifont}
\usepackage{listings}
\usepackage{multirow}
\usepackage{enumitem}
\usepackage{tabularx}
\usepackage{longtable}
\usepackage{array}
\usepackage{placeins}
\usepackage{tikz}
\usepackage{pgfplots}
\pgfplotsset{compat=1.18}

\renewcommand{\today}{Technical Report}
\pdftrailerid{redacted}
\hypersetup{
    colorlinks = true,
    citecolor = {blue},
    linkcolor = {blue},
    urlcolor = {blue},
}


\definecolor{backcolour}{RGB}{245, 245, 245}
\definecolor{commentgreen}{RGB}{34, 139, 34}
\definecolor{keywordblue}{RGB}{0, 0, 205}
\definecolor{stringred}{RGB}{178, 34, 34}
\definecolor{linenumber}{RGB}{150, 150, 150}
\definecolor{SoftBlue}{RGB}{225, 235, 248}

\lstdefinestyle{AOHPStyle}{
    backgroundcolor=\color{backcolour},
    basicstyle=\ttfamily\footnotesize,
    frame=single,
    numbers=left,
    numbersep=10pt,
    xleftmargin=15pt,
    framexleftmargin=2pt,
    language=Python,
    commentstyle=\color{commentgreen},
    stringstyle=\color{stringred},
    numberstyle=\color{linenumber},
    keywordstyle=\color{keywordblue}\bfseries,
    breakatwhitespace=false,
    breaklines=true,
    keepspaces=true,
    showspaces=false,
    showstringspaces=false,
    showtabs=false,
    tabsize=2,
}
\graphicspath{{media/}}
\date{}
\newcommand{\sys}{\textsc{AOHP}\xspace}
\newcommand{\android}{Android\xspace}
\newcommand{\aohpfull}{Android Open Harness Project\xspace}

\title{\sys: An Open-Source OS-Level Agent Harness for Personalized, Efficient and Secure Interaction}

\author{
Shanhui Zhao$^1$$^*$, Jiacheng Liu$^2$$^*$, Guohong Liu$^1$$^*$, Jichao Yan$^1$, Jialei Ye$^2$, Yuhao Yang$^3$, 
Hao Wen$^1$, Shizuo Tian$^1$, Yizhen Yuan$^1$, Yuxuan Chen$^1$, 
Yunxin Liu$^1$$^\dagger$, Ju Ren$^1$, Ya-Qin Zhang$^1$, Chao Huang$^3$, Yao Guo$^2$, Yuanchun Li$^1$$^\dagger$ \\
$^1$Tsinghua University~~~~$^2$Peking University~~~~$^3$The University of Hong Kong\\~\\
$^*$Co-primary authors.  $^\dagger$Corresponding Authors (\{liuyunxin,liyuanchun\}@air.tsinghua.edu.cn) \\
\textbf{Source Code}: \url{https://github.com/aohp-os/aohp} \\~\\
}

\begin{abstract}
AI agents are driving a new software paradigm, with the ability to autonomously call tools, extract information, manage memory, and complete tasks that span applications and data sources. Most existing end-user operating systems, however, are designed for application-centric workflows and offer little native support for AI agents. This mismatch limits the wider adoption of agents and leads to execution overhead and safety risks when running agents on conventional systems.

While the concept of agent-native operating systems is emerging, the research community lacks an open testbed to explore the architectural primitives desired for agent-mediated interaction. We present \sys (\aohpfull), an OS-level agent harness built on the Android Open Source Project (AOSP). The core design principle of \sys is to treat agents as first-class OS actors, enabling adaptive user interfaces and agent-friendly runtime environments. \sys preserves the mature Android software and hardware ecosystem while introducing three agent-oriented system mechanisms: personalized service composition, efficient agent interfaces, and secure information flow. Based on preliminary experiments on challenging tasks covering key capabilities of OS agents, \sys shows clear advantages in task completion (+21.12\% completion rate), execution cost (-51.55\% token cost), and security-policy compliance.
\end{abstract}

\begin{document}
\maketitle

\section{Introduction}

AI agents are fundamentally changing how humans interact with digital environments, moving from isolated chat interfaces into the operating system itself. Modern agents can call command-line tools, inspect graphical interfaces, use application APIs, and coordinate multi-step tasks \cite{yao2022react,openclaw,claudecode}. Agent-mediated interaction changes the role of the operating system: the OS still hosts applications, but it also needs to support agents that observe services, invoke capabilities, and enforce user intent across app boundaries.

Conventional personal operating systems are app-centric by design. Interfaces are fixed by system and application developers. GUIs are rendered for direct human visual operation rather than agent manipulation. Software lifecycle assumes one active app at a time, and permission mechanisms enforce protections at app boundaries but fail to track sensitive data across an agent's context and tool calls. These assumptions create significant execution overhead and security gaps for agent-mediated workflows.

\sys addresses this gap by redesigning \android as an agent-native harness via OS-level abstractions and framework modifications. It preserves \android's hardware support, open-source framework, and app compatibility while adding mechanisms that make services composable, efficient, and auditable for OS-level agents.

\begin{figure}[t]
\centering
\includegraphics[width=0.95\linewidth]{"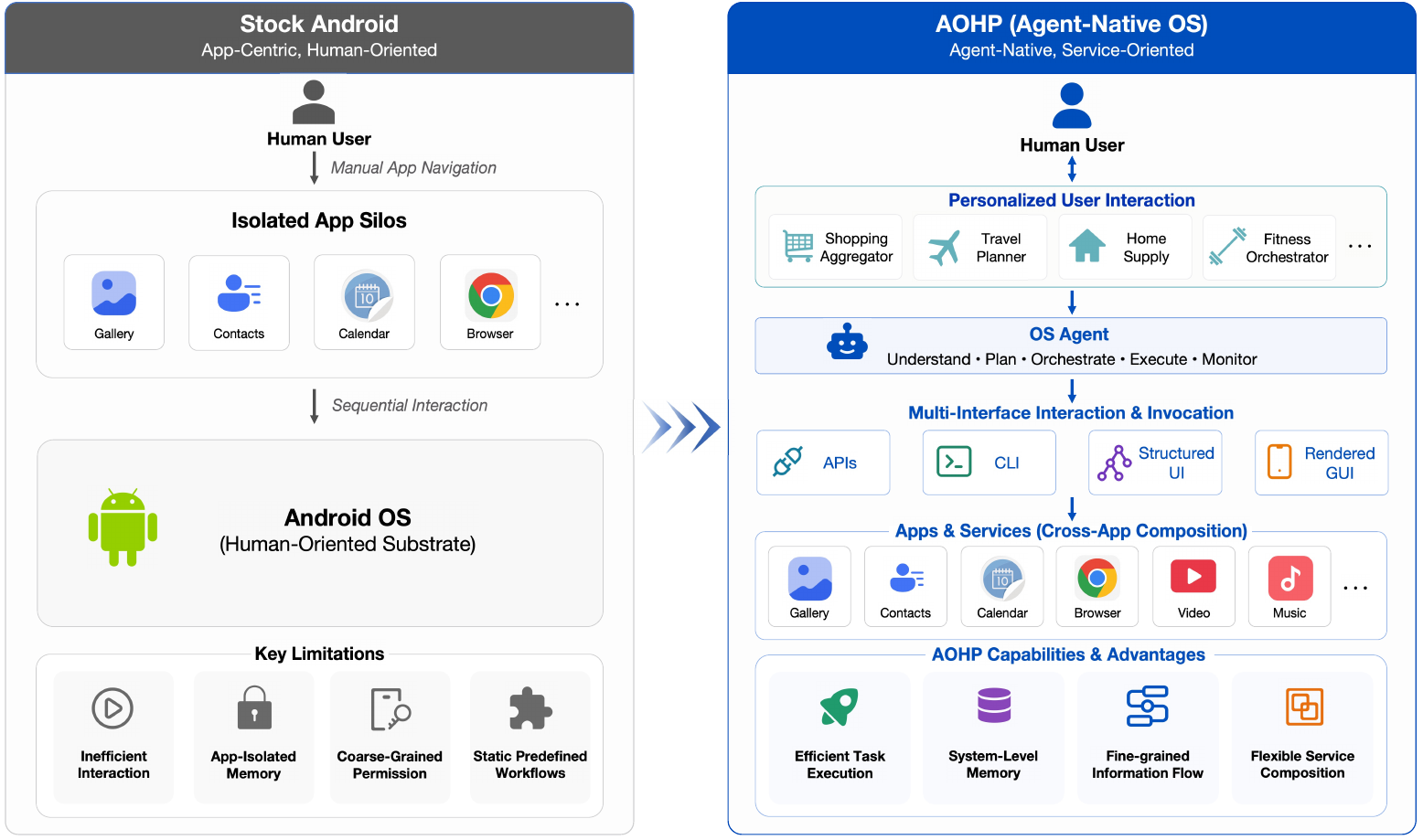"}
\caption{App-centric \android and agent-native \sys.}
\label{fig:paradigm-shift}
\end{figure}

With such OS-level supports, AOHP is able to create a new experience where \textbf{the OS can proactively adapt to the user} by generating personalized service interfaces around user intent, rather than conventionally letting users adapt to the predefined fixed OS and app interfaces. For example, instead of requiring manual switching among individual shopping apps, \sys can expose a task-level shopping entrance that aggregates product information and purchase actions across multiple services. The user interacts with the high-level concept of shopping; the OS agent handles service discovery, invocation, memory management, and policy enforcement under the hood. Figure~\ref{fig:paradigm-shift} illustrates this difference between app-centric \android and agent-native \sys.

In summary, this report makes the following contributions:
\begin{itemize}[leftmargin=1.5em]
    \item \textbf{Agent-Native OS Architecture:} We identify the architectural mismatch between app-centric OSes and agent workflows, proposing a design where services are interface-neutral capabilities, and the OS manages cross-app personalization and sensitive state.
    \item \textbf{System Implementation:} We design and implement \sys on AOSP, introducing three core mechanisms: (1) \textit{personalized service composition} for synthesizing task-level entrances; (2) \textit{efficient agent interfaces} supporting parallel background execution, structured UI, and event streams; and (3) \textit{secure information flow} that sandboxes sensitive values via information flow tracking.
    \item \textbf{Empirical Evaluation:} We evaluate \sys using OpenClaw agents on a set of self-crafted mobile tasks that demand complex cross-app interaction. Compared to stock \android, \sys raises the average completion rate from 54.44\% to 75.56\%, reduces LLM token consumption by 51.55\% and accelerates task execution by 44.21\%. Security case studies confirm that \sys effectively restricts private plaintext exposure while preserving legitimate task execution.
\end{itemize}


\section{Background \& Motivation}
\label{sec:motivation}

The need for \sys comes from a change in \emph{how} users interact with digital services.

\subsection{Emerging Demand of Adaptive User Interfaces}

Traditional operating systems expose functionality through app-defined interfaces. Users see menus, buttons, pages, and workflows chosen by the OS and app developers. This means that users can only passively adapt to the services and information determined by the developers, which may be tedious and distracting, even biasing the users' perception. The issue becomes increasingly severe as the apps are growing larger and dominating.

A promising solution to develop adaptive user interfaces. In an adaptive OS, the user interface can be deeply personalized around user intent. For example, a user who shops across multiple marketplaces should not need to reason about which app owns which capability. Instead, the OS can generate a shopping entrance that aggregates product search, comparison, coupons, delivery constraints, and purchase actions from different services. The entrance is not a static app installed by one developer; it is assembled from app, GUI, CLI, and API interfaces and personalized by system memory.

\subsection{Agents Become a Major Workload}

AI agents are promising to enable such adaptive experiences.
Modern agents can operate applications, issue commands, and interact with web or app services. In many workflows, the agent can serve as an execution layer between user intent and app-level operations: it reads application state, invokes system functions, transfers data across services, waits for asynchronous events, and acts repeatedly over time.

Agent-mediated interaction exposes a gap in conventional OS design. Existing systems assume app-mediated interaction, while agents have different operational properties. They process structured text more efficiently than pixels, run multiple tasks in parallel, retain long-lived context, and make tool calls faster than users can inspect them. A system designed around app-mediated interaction cannot make agent execution efficient or safe by construction.

\subsection{Why We Need A OS-level Harness}

To make the agent-mediated adaptive user interfaces feasible, efficient and secure, many components in the system and framework layers should be redesigned, while many other componnents can be reused. Therefore, building a harness over an existing mature OS is an ideal choice.
We choose \android as the substrate for \sys for practical reasons.

\textbf{Rich application ecosystem.} \android offers a broad app ecosystem covering communication, productivity, commerce, content, and device control, which is useful to foster rich custom services.

\textbf{Mature hardware support.} \android already runs on a wide spectrum of devices with established support for drivers, sensors, networking, and power management.

\textbf{Open-source accessibility.} The Android Open Source Project allows deep modifications to system services, framework layers, UI stacks, and runtime policies, which makes it a practical base for fine-grained systems research \cite{androidaosp}.



\section{System Design}
\label{sec:system-design}

\sys turns the requirements above into three design principles. First, \textbf{user interfaces should be composed on demand}: the OS agent should break app and service silos by discovering and recomposing capabilities across providers into personalized interaction surfaces driven by the user's intent, context, and state. Second, \textbf{agent-service interaction should be compatible, efficient, and interface-neutral}: the OS should provide a unified substrate that preserves legacy \android app interaction while supporting agent-oriented service logic, so agents can access capabilities with lower overhead and higher accuracy. Third, \textbf{sensitive data should be isolated by default}: agents should not observe private plaintext by default, while the system enforces fine-grained information-flow tracking, approval, and audit throughout the task.

\begin{figure*}[t]
\centering
\includegraphics[width=\textwidth]{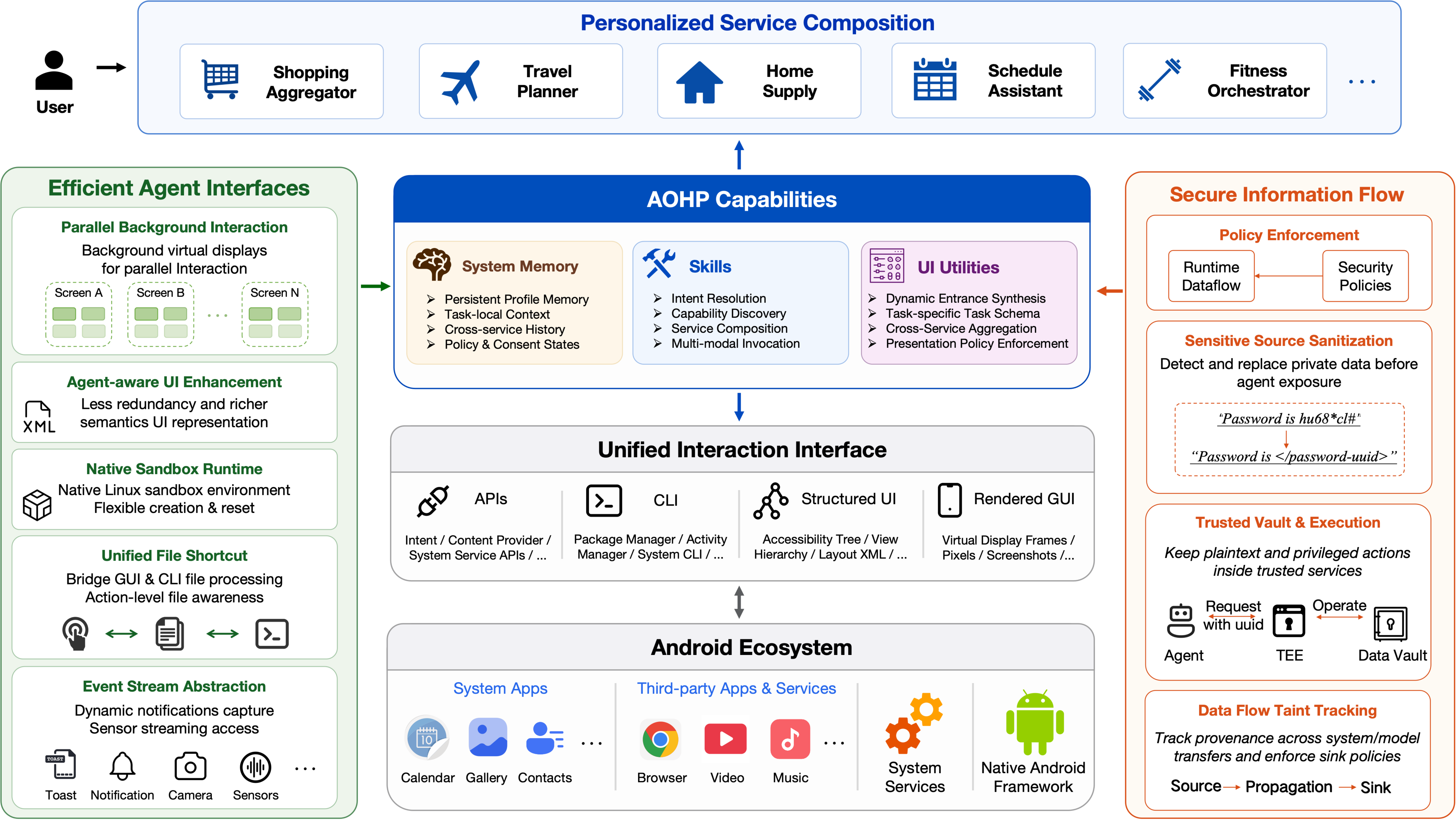}
\caption{Architecture overview of \sys.}
\label{fig:aohp-architecture}
\end{figure*}

Figure~\ref{fig:aohp-architecture} shows the resulting architecture. Vertically, \sys is organized into four layers. The bottom layer is the \textbf{\android ecosystem}, which keeps existing apps, system services, hardware resources, and platform APIs as the compatibility base. Above it, the \textbf{unified interaction interface} normalizes both traditional \android interfaces and emerging agent interfaces into four invocation modes: API, CLI, Structured UI, and Rendered GUI. This layer lets agents select a compact symbolic path when available and fall back to visual operation when compatibility requires it.

The third layer is the \textbf{\sys capabilities} layer. It reorganizes services and agent functions provided by apps, system components, and agent tools into system memory, skills, and UI utilities. System memory stores preferences, task state, histories, and policy decisions outside any single app. Skills package reusable service capabilities and execution routines. UI utilities support the construction of generated entrances and task-specific interaction surfaces. The top layer is \textbf{personalized service composition}, where these capabilities are assembled into new user-facing application surfaces tailored to the user's current task. This layer represents the application model enabled by \sys: the user interacts with a personalized service entrance, while the OS agent resolves the underlying service graph and execution path.

Horizontally, \sys includes two cross-layer mechanisms. \textbf{Efficient agent interfaces} optimize how agents access system resources, app state, files, events, and isolated execution substrates. They reduce visual-processing overhead, avoid unnecessary foreground serialization, and make agent actions less dependent on brittle GUI navigation. \textbf{Secure information flow} provides the stricter protection model required by agent-mediated execution. It sanitizes sensitive sources, routes private values through trusted vault operations, propagates taint metadata, and records auditable traces before sensitive data reaches external sinks or state-changing actions.

The following sections detail the three core parts of this design. Section~\ref{sec:personalized-service-composition} explains how \sys discovers and recomposes underlying capabilities into personalized service entrances. Section~\ref{sec:efficient-interfaces} describes the agent-native interfaces that make service access and task execution efficient. Section~\ref{sec:secure-flow} presents the security mechanisms that isolate private data and enforce system-level information-flow policies.

\section{Personalized Service Composition}
\label{sec:personalized-service-composition}

The most visible change in \sys is that interaction surfaces become personalized and generated rather than fully predefined. A conventional app exposes developer-chosen functions. \sys lets the OS agent synthesize service entrances around recurring goals, shifting interaction from app navigation to task-level service access.

\subsection{Generated Service Entrances}

Generated entrances are user-facing shells backed by OS-managed service composition. A shopping entrance can aggregate product search from multiple providers, normalize product attributes, apply preferences such as size and budget, and expose a task-specific interface for comparison and purchase.

Each entrance contains three parts: a task schema, a service graph, and a presentation policy. The task schema defines what the user is trying to accomplish, such as ``compare shoes under a budget'' or ``refill household supplies''. The service graph maps this task to concrete service capabilities. The presentation policy decides which intermediate results should be shown to the user and which can remain agent-internal. This separation lets \sys personalize the entrance without hiding consequential decisions.

\subsection{Capability Discovery and Composition}

\sys builds these entrances by discovering service capabilities across API, CLI, and GUI channels. Each capability is represented with input/output schemas, preconditions, side effects, and policy labels. The OS agent can then compose capabilities into a higher-level workflow. This design lets legacy apps participate through GUI export while allowing newer services to expose more direct APIs.

Composition is constrained by policy. For example, product search may be freely parallelized across providers, whereas purchase submission is a state-changing action that requires explicit confirmation. Similarly, a delivery address can be used to estimate shipping only through the information-flow sandbox. The generated entrance is both a convenience layer and a policy-enforcement surface.

\subsection{Cross-Service Personalization}

System memory lets personalization survive app boundaries. Preferences learned while using one service can improve another service, subject to policy. For example, a user's preferred delivery window learned from one shopping workflow can be used when comparing products from another marketplace. The design requirement is that memory remains OS-mediated: personalization can be shared, audited, and revoked without depending on each app's private data model.

\sys distinguishes between persistent profile memory, task-local memory, and sensitive memory. Persistent profile memory stores stable preferences. Task-local memory stores temporary state, such as candidate products or partially completed forms. Sensitive memory stores private values through sandbox indices. This distinction prevents personalization from becoming an uncontrolled accumulation of private context.

\section{Efficient Agent Interfaces}
\label{sec:efficient-interfaces}

\sys exposes execution environments, UI semantics, storage, and events as agent-native primitives. These abstractions reduce visual-processing overhead, rigid sequential execution, and brittle cross-app handoffs in agent-mediated workflows.

\subsection{Parallel Background Interaction}
Conventional mobile OSes couple app lifecycles to the physical display. \sys decouples execution from the screen through lightweight virtual displays, allowing agents to run waiting-heavy or independent workflows in the background without preempting the active foreground session.

\subsection{Agent-aware UI Enhancement}
Conventional app GUIs contain rendering details that are unnecessary for agent reasoning. \sys abstracts GUIs into structured representations with lower redundancy and richer semantics, while retaining rendered GUI fallback for visual components.

\subsection{Native Sandbox Runtime}
App-mediated GUI, API, and CLI paths do not cover all agent work. Agents often need a local execution substrate for computation, transformation, and tooling. \sys includes a native, OS-managed sandbox runtime that can be created and reset as an execution surface independent of app-facing interfaces. Agents can execute code, process data, and host long-running services inside the sandbox, then return structured results to the task context without placing intermediate steps in the agent context.

\subsection{Unified File Shortcut}
Cross-app agent workflows often rely on files as shared intermediate artifacts, such as saving an attachment in one app and reusing it in another. On stock systems, these artifacts remain implicit: GUI actions may create or modify storage, but agents lack a stable, OS-level account of the outcome. Conversely, programmatic file operations have no uniform way to invoke app-native affordances such as system share flows when the destination expects them.
\sys treats files as first-class task objects at the OS boundary. GUI interactions that affect storage are reflected back as structured file observations, so agents can reason about what changed without inferring paths from screenshots or per-app pickers. In the reverse direction, the same layer lets agents hand a resolved artifact to another interface, either through direct programmatic access or by launching the appropriate system UI flow on a chosen display. This unifies file-producing GUI steps and file-consuming programmatic steps into a single cross-app data plane and reduces brittle handoffs across opaque storage layouts.

\subsection{Event Stream Abstraction}
Operating systems continuously generate asynchronous and transient events that are difficult to capture with request-response interfaces. \sys introduces an \textit{Event Stream} abstraction that lets agents subscribe to, process, and unsubscribe from continuous data sources. It currently supports two stream types:
\begin{itemize}
    \item \textbf{Dynamic Notification Capture:} Transient system events, such as Toasts, pop-ups, or push notifications, often disappear before an agent can poll them. \sys implements a notification buffer to intercept and retain these short-lived messages, so agents do not miss critical UI context.
    \item \textbf{Sensor Streaming Access:} To perceive the physical environment, \sys streams hardware sensor data (e.g., accelerometer, gyroscope, microphone, or camera events). Agents can process real-time physical states without repeated polling.
\end{itemize}
The abstraction separates event generation from consumption, allowing agents to react to system and environmental changes without repeated polling.

\section{Secure Information Flow}
\label{sec:secure-flow}

\sys treats sensitive data as OS-controlled state rather than agent-visible context. By default, private plaintext is replaced with typed references before it reaches the agent; trusted system components mediate plaintext operations, external transfers, and approvals while preserving evidence for audit. This model is necessary because agent tasks cross app, tool, memory, and service boundaries where conventional app permissions cannot track how private data propagates.

\subsection{Policy Enforcement}

\sys enforces privacy policies over runtime data use rather than only over static permissions or application identities. For each sensitive operation, the policy layer evaluates the data source, requested purpose, destination, action sensitivity, and approval state. This lets ordinary non-sensitive flows proceed while requiring consent or blocking transfers and state-changing actions involving private data.

The same policy context makes authorization more understandable to users. When approval is needed, \sys can explain the requested use in terms of source, purpose, destination, and downstream effect, rather than presenting an opaque permission prompt. Thus, enforcement is tied to the concrete use of private data in the task.

\subsection{Sensitive Source Sanitization}

\sys adopts a conservative protection strategy for sensitive sources. Before sensitive content enters the agent context, \sys replaces plaintext with typed placeholders such as \texttt{<payment-card:uuid>} or \texttt{<home-address:uuid>}. These placeholders preserve task-level meaning while hiding the underlying value.

The system maintains a data vault that stores sensitive values behind opaque identifiers. \sys sanitizes supported sources, including application pages, files, event streams, API responses, system memory, and user interactions. Developer-provided annotations can mark sensitive fields explicitly; when annotations are unavailable, \sys applies conservative detection rules to protect likely sensitive content.

\subsection{Trusted Vault and Execution}

When an agent needs to operate on sensitive information, it submits an intent to a trusted vault executor using the corresponding placeholder. The executor checks policy, obtains user approval when necessary, and performs operations such as formatting, comparison, validation, or composition inside the trusted environment. If the result is still sensitive, the executor returns another placeholder rather than plaintext.

The trusted executor also mediates sensitive transfers to external interfaces. For GUI use, it can fill an approved field directly; for API or CLI use, it can substitute plaintext at the system boundary while keeping it out of the agent context. This design lets agents complete tasks that require private data while exposing sensitive values only to trusted system components.

\subsection{Data-Flow Taint Tracking}

Sanitization protects sensitive sources at entry points; taint tracking preserves their provenance after they are used. Once sensitive data enters \sys, it is associated with taint metadata that follows the value through copying, transformation, composition, and transfer. This allows \sys to preserve the information-flow chain even when private data is used indirectly across multiple task steps.

At system exits and other policy-relevant boundaries, \sys checks tainted data before it is displayed, stored, submitted, or transmitted. The resulting taint path also provides an audit trail for explaining which source reached which sink through which task steps. In this way, the agent can reason over sensitive references, while the operating system remains responsible for tracking and controlling when private data may leave trusted components.

\section{Evaluation}
\label{sec:evaluation}

We evaluate \sys with the OpenClaw agent and compare it against stock \android. Stock \android is the app-centric baseline: the agent receives conventional GUI observations and operates through ordinary app interactions. \sys is the agent-native setting: the same agent additionally uses generated service entrances, structured UI observations, system CLIs/APIs, virtual execution, and sandboxed information flow.

Our benchmark targets the capabilities required for \emph{personalized service composition}. It contains 30 real-world mobile tasks built exclusively from real applications. The tasks span five core capability categories: (i) GUI operation, (ii) non-GUI operation, (iii) event capture, (iv) multi-source information retrieval, and (v) memory management. A sixth hybrid category composes these primitives into longer cross-capability workflows, mirroring the service-composition setting that \sys is designed for. Each category contains five tasks; Appendix~\ref{app:benchmark-tasks} lists all of them. We measure task completion, success count, tool calls, duration, token usage, LLM requests, and security-policy behavior.

\subsection{Functionality and Performance}

\begin{figure*}[t]
\centering
\includegraphics[width=\textwidth]{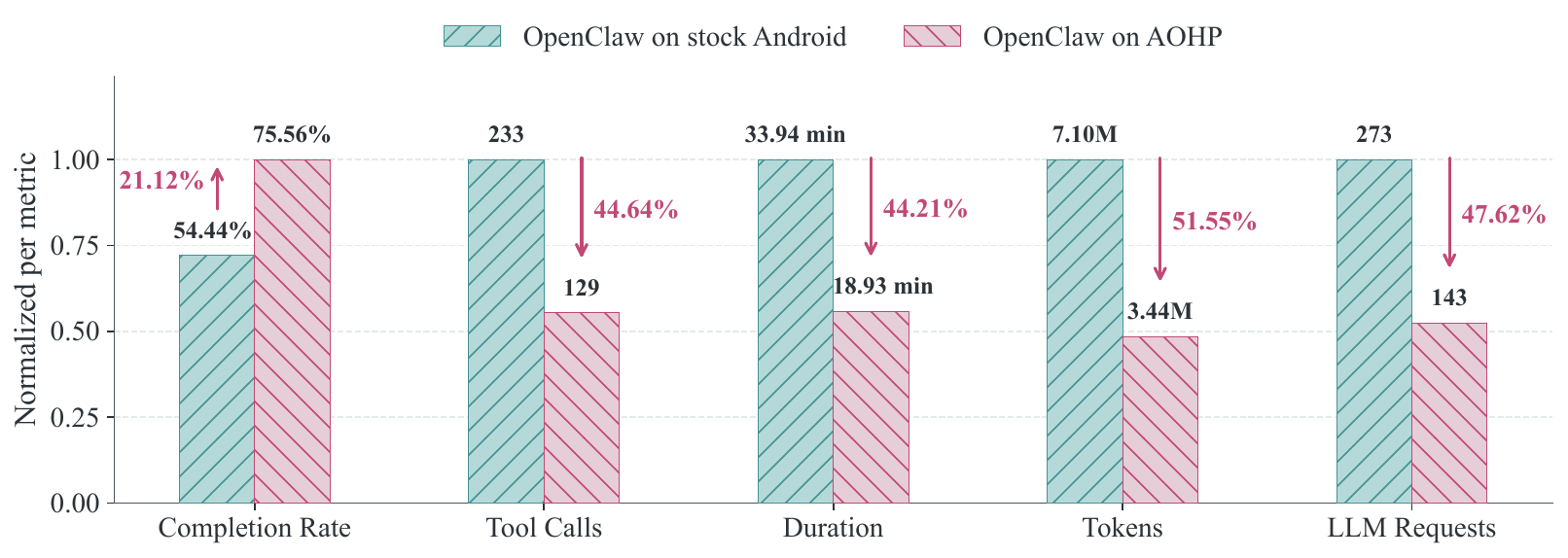}
\caption{Overall comparison between OpenClaw on \sys and stock \android. Completion rate is averaged over all tasks; tool calls, duration, tokens, and LLM requests are aggregated over the tasks both settings solve completely.}
\label{fig:overall-results}
\end{figure*}

We run all benchmark tasks with the OpenClaw agent under both settings. Each task is scored at the granularity of objective checkpoints, so a task that is only partially completed still earns partial credit; we report the resulting checkpoint-weighted \emph{completion rate}. Figure~\ref{fig:overall-results} summarizes the outcome.

On \sys, OpenClaw reaches a 75.56\% average completion rate (20 tasks fully solved and 5 tasks partially completed). With the same agent on stock \android, the completion rate drops to 54.44\% (13 tasks fully solved and 7 tasks partially completed). \sys thus raises the average completion rate by 21.12 points and fully solves seven more tasks than the baseline. The gains concentrate on tasks involving transient notifications, fine-grained in-app GUI manipulation, and multi-step cross-app or memory-dependent workflows, where \sys leverages the interfaces provided in Section~\ref{sec:efficient-interfaces} to improve the capabilities of the agent.

\subsection{Efficiency Analysis}

To compare execution cost on equal footing, we restrict the efficiency analysis to the 11 tasks that both settings solve completely. This removes the confound of unequal task difficulty and isolates the cost of producing the same successful outcome. Table~\ref{tab:efficiency} reports the aggregates over this commonly-solved subset, and we analyze the two cost dimensions below.

\begin{table}[t]
\centering
\small
\caption{Execution cost on the tasks solved completely by both settings.}
\label{tab:efficiency}
\begingroup
\setlength{\tabcolsep}{5pt}
\begin{tabularx}{0.90\linewidth}{@{}l *{4}{>{\centering\arraybackslash}X}@{}}
\toprule
\textbf{Setting} & \textbf{Tool Calls} & \textbf{Duration (min)} & \textbf{Tokens} & \textbf{LLM Requests} \\
\midrule
OpenClaw on stock \android & 233 & 33.94 & 7,103,192 & 273 \\
OpenClaw on \sys & \textbf{129} & \textbf{18.93} & \textbf{3,441,759} & \textbf{143} \\
\midrule
Reduction & 44.64\% & 44.21\% & 51.55\% & 47.62\% \\
\bottomrule
\end{tabularx}
\endgroup
\end{table}

\textbf{Tool Calls and Duration.} Even though the baseline agent can drive some operations through direct shell commands, stock \android offers few agent-native interfaces, so cross-app workflows and in-app GUI operations still rely largely on GUI navigation---climbing view hierarchies, scrolling, and re-issuing taps as interface depth grows. \sys shortens these paths through the \textit{Unified File Shortcut}, \textit{Agent-aware UI Enhancement}, and \textit{Event Stream Abstraction}, which offer more agent-oriented interaction. The savings therefore come from interface-intensive and data-processing workflows rather than a uniform per-step speedup, yielding a 44.64\% reduction in tool calls and a 44.21\% reduction in total duration across the commonly-solved subset.

\textbf{Token Consumption and LLM Requests.} Token cost grows with the number of steps, since the tool calls and observations accumulated during execution keep enlarging the context. Because \sys completes the same tasks in fewer steps and returns more compact observations, both prompt length and the number of round-trips shrink. On the commonly-solved subset, \sys cuts total tokens by 51.55\% (3.44M vs.\ 7.10M) and LLM requests by 47.62\% (143 vs.\ 273), with input tokens down 51.50\% and output tokens down 57.48\%.

\subsection{Information-Flow Security}

\sys evaluates information-flow security with a purpose-built annotated payment application. The app exposes account information, payment credentials, transfer inputs, confirmation actions, invoice files, and transaction-related event streams. Around this app, we construct case-driven tests that exercise source sanitization, sink and action mediation, vault-token use, taint propagation, file handling, event handling, and fail-closed behavior. Table~\ref{tab:security-cases} summarizes the main enforced cases. In our prototype evaluation on this annotated payment app, \sys enforced all five cases in Table~\ref{tab:security-cases} as expected.

\begin{table}[!thbp]
\centering
\small
\caption{Security checks using an annotated payment application.}
\label{tab:security-cases}
\begin{tabularx}{\linewidth}{p{2.6cm} X p{1.7cm}}
\toprule
\textbf{Check} & \textbf{Expected Behavior} & \textbf{Result} \\
\midrule
Sensitive display & Account, card, phone, and transaction fields appear as vault references in the agent-visible UI, not plaintext. & Pass \\
Ordinary actions & Non-sensitive controls and ordinary files are allowed without extra approval. & Pass \\
Sensitive actions & Transfer fields, payment confirmation, and sensitive file sharing require user consent. & Pass \\
Unsupported access & Requests outside the declared policy scope fail closed instead of exposing sensitive data. & Pass \\
Sensitive events & Transaction-related event streams redact sensitive fields and preserve taint metadata. & Pass \\
\bottomrule
\end{tabularx}
\end{table}
\FloatBarrier

\section{Related Work}
\label{sec:related}

\subsection{GUI Agents and Mobile Task Automation}

Recent work has shown that large-language-model-based agents can operate computers and smartphones through graphical interfaces. Existing approaches span modular systems and end-to-end policies. Modular systems such as the AutoDroid series, Agent S, and OS-Copilot compose planners, retrievers, memory modules, and grounding components to improve reuse and error recovery \cite{autodroid,autodroidv2,agashe2025agent,os-copilot}. End-to-end approaches such as AppAgent, UI-TARS, Mobile-Agent, and SeeAct directly infer actions from current observations, often using multimodal models over screenshots or UI descriptions \cite{zhang2025appagent,uitars,zheng2024gpt,wang2024mobile,mobileagentv3}. Related benchmarks, such as AndroidWorld, OSWorld, AndroidLab, and MobileWorld, evaluate agents in dynamic GUI environments \cite{rawles2025androidworld,xie2024osworld,xu2025androidlab,kong2025mobileworld}.

These systems improve GUI-agent performance, but they mainly treat the operating system as a fixed substrate. \sys studies the execution environment itself: instead of proposing only a stronger agent policy, it redesigns the OS interfaces and enforcement mechanisms that such policies depend on.

\subsection{Android Automation and Interface Abstraction}

Before the rise of LLM agents, Android automation and testing systems had already explored how to interact with mobile interfaces programmatically. Systems such as DroidBot emphasized UI-guided exploration, while earlier instruction-following work mapped natural language requests into mobile action sequences \cite{li2017droidbot,seq2act}. More recent interface-understanding models, including ScreenAI, OS-ATLAS, Aguvis, and other GUI grounding methods, improve semantic interpretation of screen content \cite{baechler2024screenai,wu2025atlas,xu2024aguvis,cheng2024seeclick}.

\sys uses screen understanding as one component. Existing automation and GUI-agent systems still inherit app-centric rendering paths, foreground-oriented execution assumptions, and app-defined interaction surfaces. \sys moves these concerns into operating-system abstractions, including personalized service entrances, agent-aware UI enhancement, parallel background interaction, and system-managed task traces.

\subsection{System Safety, Provenance, and Recoverability}

Recent security studies show that tool-using agents face a broad attack surface in which untrusted external content can hijack tool execution or exfiltrate private data. Indirect prompt injection first exposed how LLM-integrated applications blur the boundary between data and instructions \cite{greshake2023not}; benchmarks such as InjecAgent and AgentDojo further systematize these risks in realistic tool-use settings \cite{zhan2024injecagent,debenedetti2024agentdojo}. These results motivate system mechanisms that mediate data, authority, and actions rather than relying only on model-level robustness.

Recent systems defenses have moved toward deterministic enforcement at execution boundaries. FIDES and f-secure LLM systems apply information-flow control to agent planners and LLM pipelines, using labels, taint propagation, and policy checks to prevent untrusted data from steering sensitive decisions \cite{costa2025securing,wu2024system}. \sys follows this systems-oriented direction but moves the enforcement boundary to the mobile OS. Building on the mobile taint-tracking lineage of TaintDroid \cite{enck2014taintdroid}, \sys tracks and mediates sensitive data across app UI, files, event streams, vault references, generated entrances, and user approval loops.

\section{Conclusion and Future Work}
\label{sec:conclusion}

\sys is an agent-native open fork of \android for agent-mediated personal computing. It keeps Android compatibility while adding personalized service composition, efficient agent interfaces, and secure information flow. Preliminary results with OpenClaw show higher task completion and lower execution cost than stock \android; the security case study shows that sensitive data can be redacted, tracked, and mediated at system boundaries.

\textbf{Compatibility coverage.} Future work should expand support for apps with custom rendering, anti-automation logic, and undocumented behavior. This requires stronger structured UI extraction, rendered-GUI fallback that fails predictably, and clearer compatibility guidelines for app developers.

\textbf{Capability discovery.} \sys depends on accurate service descriptions, side-effect labels, and policy metadata. Future versions should combine developer-provided descriptors with automatic capability inference so that legacy apps can be integrated with less manual annotation.

\textbf{Resource scheduling.} Background execution must respect mobile resource, thermal, and memory limits. A fuller prototype should expose scheduling policies for virtual displays, sandbox runtimes, event streams, and foreground user activity.

\textbf{Policy usability.} Fine-grained information-flow control depends on approvals that users can understand without excessive interruption. Future work should improve approval UI, trace review, and policy explanations for purpose, destination, retention, and consent.

\bibliographystyle{plain}
\bibliography{references}

@article{yao2022react,
  title={React: Synergizing reasoning and acting in language models},
  author={Yao, Shunyu and Zhao, Jeffrey and Yu, Dian and Du, Nan and Shafran, Izhak and Narasimhan, Karthik and Cao, Yuan},
  journal={arXiv preprint arXiv:2210.03629},
  year={2022}
}

@misc{openclaw,
  title={{OpenClaw}},
  author={{OpenClaw Contributors}},
  year={2026},
  howpublished={\url{https://docs.openclaw.ai/}},
  note={Accessed: 2026-06-11}
}

@misc{claudecode,
  title={{Claude Code}},
  author={{Anthropic}},
  year={2026},
  howpublished={\url{https://docs.anthropic.com/en/docs/claude-code/overview}},
  note={Accessed: 2026-06-11}
}

@misc{androidaosp,
  title={Android Open Source Project},
  author={{Google}},
  year={2026},
  howpublished={\url{https://source.android.com/}}
}

@inproceedings{greshake2023not,
  title={Not what you've signed up for: Compromising real-world llm-integrated applications with indirect prompt injection},
  author={Greshake, Kai and Abdelnabi, Sahar and Mishra, Shailesh and Endres, Christoph and Holz, Thorsten and Fritz, Mario},
  booktitle={Proceedings of the 16th ACM workshop on artificial intelligence and security},
  pages={79--90},
  year={2023}
}

@inproceedings{zhan2024injecagent,
  title={Injecagent: Benchmarking indirect prompt injections in tool-integrated large language model agents},
  author={Zhan, Qiusi and Liang, Zhixiang and Ying, Zifan and Kang, Daniel},
  booktitle={Findings of the Association for Computational Linguistics: ACL 2024},
  pages={10471--10506},
  year={2024}
}

@article{debenedetti2024agentdojo,
  title={Agentdojo: A dynamic environment to evaluate prompt injection attacks and defenses for llm agents},
  author={Debenedetti, Edoardo and Zhang, Jie and Balunovic, Mislav and Beurer-Kellner, Luca and Fischer, Marc and Tram{\`e}r, Florian},
  journal={Advances in Neural Information Processing Systems},
  volume={37},
  pages={82895--82920},
  year={2024}
}

@article{enck2014taintdroid,
  title={Taintdroid: an information-flow tracking system for realtime privacy monitoring on smartphones},
  author={Enck, William and Gilbert, Peter and Han, Seungyeop and Tendulkar, Vasant and Chun, Byung-Gon and Cox, Landon P and Jung, Jaeyeon and McDaniel, Patrick and Sheth, Anmol N},
  journal={ACM Transactions on Computer Systems (TOCS)},
  volume={32},
  number={2},
  pages={1--29},
  year={2014},
  publisher={ACM New York, NY, USA}
}

@article{costa2025securing,
  title={Securing ai agents with information-flow control},
  author={Costa, Manuel and K{\"o}pf, Boris and Kolluri, Aashish and Paverd, Andrew and Russinovich, Mark and Salem, Ahmed and Tople, Shruti and Wutschitz, Lukas and Zanella-B{\'e}guelin, Santiago},
  journal={arXiv preprint arXiv:2505.23643},
  year={2025}
}

@article{wu2024system,
  title={System-level defense against indirect prompt injection attacks: An information flow control perspective},
  author={Wu, Fangzhou and Cecchetti, Ethan and Xiao, Chaowei},
  journal={arXiv preprint arXiv:2409.19091},
  year={2024}
}

@inproceedings{seq2act,
  title={Mapping natural language instructions to mobile UI action sequences},
  author={Li, Yang and He, Jiacong and Zhou, Xin and Zhang, Yuan and Baldridge, Jason},
  booktitle={Proceedings of the 58th annual meeting of the association for computational linguistics},
  pages={8198--8210},
  year={2020}
}

@inproceedings{li2017droidbot,
  title={Droidbot: a lightweight ui-guided test input generator for android},
  author={Li, Yuanchun and Yang, Ziyue and Guo, Yao and Chen, Xiangqun},
  booktitle={2017 IEEE/ACM 39th international conference on software engineering companion (ICSE-C)},
  pages={23--26},
  year={2017},
  organization={IEEE}
}

@inproceedings{rawles2025androidworld,
  title={Androidworld: A dynamic benchmarking environment for autonomous agents},
  author={Rawles, Chris and Clinckemaillie, Sarah and Chang, Yifan and Waltz, Jonathan and Lau, Gabrielle and Fair, Marybeth and Li, Alice and Bishop, William and Li, Wei and Campbell-Ajala, Folawiyo and others},
  booktitle={International Conference on Learning Representations},
  volume={2025},
  pages={406--441},
  year={2025}
}

@inproceedings{xu2025androidlab,
  title={Androidlab: Training and systematic benchmarking of android autonomous agents},
  author={Xu, Yifan and Liu, Xiao and Sun, Xueqiao and Cheng, Siyi and Yu, Hao and Lai, Hanyu and Zhang, Shudan and Zhang, Dan and Tang, Jie and Dong, Yuxiao},
  booktitle={Proceedings of the 63rd Annual Meeting of the Association for Computational Linguistics (Volume 1: Long Papers)},
  pages={2144--2166},
  year={2025}
}

@inproceedings{autodroidv2,
  title={Autodroid-v2: Boosting slm-based gui agents via code generation},
  author={Wen, Hao and Tian, Shizuo and Pavlov, Borislav and Du, Wenjie and Li, Yixuan and Chang, Ge and Zhao, Shanhui and Liu, Jiacheng and Liu, Yunxin and Zhang, Ya-Qin and others},
  booktitle={Proceedings of the 23rd Annual International Conference on Mobile Systems, Applications and Services},
  pages={223--235},
  year={2025}
}

@article{mobileagentv3,
  title={Mobile-agent-v3: Fundamental agents for gui automation},
  author={Ye, Jiabo and Zhang, Xi and Xu, Haiyang and Liu, Haowei and Wang, Junyang and Zhu, Zhaoqing and Zheng, Ziwei and Gao, Feiyu and Cao, Junjie and Lu, Zhengxi and others},
  journal={arXiv preprint arXiv:2508.15144},
  year={2025}
}

@article{uitars,
  title={Ui-tars: Pioneering automated gui interaction with native agents},
  author={Qin, Yujia and Ye, Yining and Fang, Junjie and Wang, Haoming and Liang, Shihao and Tian, Shizuo and Zhang, Junda and Li, Jiahao and Li, Yunxin and Huang, Shijue and others},
  journal={arXiv preprint arXiv:2501.12326},
  year={2025}
}

@inproceedings{wu2025atlas,
  title={OS-ATLAS: Foundation action model for generalist GUI agents},
  author={Wu, Zhiyong and Wu, Zhenyu and Xu, Fangzhi and Wang, Yian and Sun, Qiushi and Jia, Chengyou and Cheng, Kanzhi and Ding, Zichen and Chen, Liheng and Liang, Paul Pu and others},
  booktitle={International Conference on Learning Representations},
  volume={2025},
  pages={5090--5108},
  year={2025}
}

@article{xie2024osworld,
  title={Osworld: Benchmarking multimodal agents for open-ended tasks in real computer environments},
  author={Xie, Tianbao and Zhang, Danyang and Chen, Jixuan and Li, Xiaochuan and Zhao, Siheng and Cao, Ruisheng and Hua, Toh J and Cheng, Zhoujun and Shin, Dongchan and Lei, Fangyu and others},
  journal={Advances in Neural Information Processing Systems},
  volume={37},
  pages={52040--52094},
  year={2024}
}

@article{xu2024aguvis,
  title={Aguvis: Unified pure vision agents for autonomous gui interaction},
  author={Xu, Yiheng and Wang, Zekun and Wang, Junli and Lu, Dunjie and Xie, Tianbao and Saha, Amrita and Sahoo, Doyen and Yu, Tao and Xiong, Caiming},
  journal={arXiv preprint arXiv:2412.04454},
  year={2024}
}

@article{kong2025mobileworld,
  title={MobileWorld: Benchmarking Autonomous Mobile Agents in Agent-User Interactive and MCP-Augmented Environments},
  author={Kong, Quyu and Zhang, Xu and Yang, Zhenyu and Gao, Nolan and Liu, Chen and Tong, Panrong and Cai, Chenglin and Zhou, Hanzhang and Zhang, Jianan and Chen, Liangyu and others},
  journal={arXiv preprint arXiv:2512.19432},
  year={2025}
}

@inproceedings{autodroid,
  author = {Wen, Hao and Li, Yuanchun and Liu, Guohong and Zhao, Shanhui and Yu, Tao and Li, Toby Jia-Jun and Jiang, Shiqi and Liu, Yunhao and Zhang, Yaqin and Liu, Yunxin},
  title = {AutoDroid: LLM-powered Task Automation in Android},
  booktitle = {Proceedings of the 30th Annual International Conference on Mobile Computing and Networking},
  pages = {543--557},
  year = {2024}
}

@inproceedings{agashe2025agent,
  title={Agent s: An open agentic framework that uses computers like a human},
  author={Agashe, Saaket and Han, Jiuzhou and Gan, Shuyu and Yang, Jiachen and Li, Ang and Wang, Xin},
  booktitle={International Conference on Learning Representations},
  volume={2025},
  pages={22924--22946},
  year={2025}
}

@misc{os-copilot,
  title={OS-Copilot: Towards Generalist Computer Agents with Self-Improvement},
  author={Wu, Zhiyong and Han, Chengcheng and Ding, Zichen and Weng, Zhenmin and Liu, Zhoumianze and Yao, Shunyu and Yu, Tao and Kong, Lingpeng},
  year={2024},
  eprint={2402.07456},
  archivePrefix={arXiv},
  primaryClass={cs.AI}
}

@inproceedings{zhang2025appagent,
  title={Appagent: Multimodal agents as smartphone users},
  author={Zhang, Chi and Yang, Zhao and Liu, Jiaxuan and Li, Yanda and Han, Yucheng and Chen, Xin and Huang, Zebiao and Fu, Bin and Yu, Gang},
  booktitle={Proceedings of the 2025 CHI Conference on Human Factors in Computing Systems},
  pages={1--20},
  year={2025}
}

@article{wang2024mobile,
  title={Mobile-agent-v2: Mobile device operation assistant with effective navigation via multi-agent collaboration},
  author={Wang, Junyang and Xu, Haiyang and Jia, Haitao and Zhang, Xi and Yan, Ming and Shen, Weizhou and Zhang, Ji and Huang, Fei and Sang, Jitao},
  journal={Advances in Neural Information Processing Systems},
  volume={37},
  pages={2686--2710},
  year={2024}
}

@article{zheng2024gpt,
  title={Gpt-4v (ision) is a generalist web agent, if grounded},
  author={Zheng, Boyuan and Gou, Boyu and Kil, Jihyung and Sun, Huan and Su, Yu},
  journal={arXiv preprint arXiv:2401.01614},
  year={2024}
}

@article{baechler2024screenai,
  title={Screenai: A vision-language model for ui and infographics understanding},
  author={Baechler, Gilles and Sunkara, Srinivas and Wang, Maria and Zubach, Fedir and Mansoor, Hassan and Etter, Vincent and C{\u{a}}rbune, Victor and Lin, Jason and Chen, Jindong and Sharma, Abhanshu},
  journal={arXiv preprint arXiv:2402.04615},
  year={2024}
}

@inproceedings{cheng2024seeclick,
  title={Seeclick: Harnessing gui grounding for advanced visual gui agents},
  author={Cheng, Kanzhi and Sun, Qiushi and Chu, Yougang and Xu, Fangzhi and YanTao, Li and Zhang, Jianbing and Wu, Zhiyong},
  booktitle={Proceedings of the 62nd Annual Meeting of the Association for Computational Linguistics (Volume 1: Long Papers)},
  pages={9313--9332},
  year={2024}
}

\appendix
\onecolumn

\section{Benchmark Tasks}
\label{app:benchmark-tasks}

\small
\newcounter{benchtask}
\newcommand{\benchcat}[1]{%
  \midrule
  \setcounter{benchtask}{0}%
  \textbf{\textit{#1}}\\[0.4em]%
}
\newcommand{\benchtask}[1]{%
  \stepcounter{benchtask}%
  \noindent\hangafter=1\hangindent=1.6em\textbf{\arabic{benchtask}.}~#1\\[0.55em]%
}
\newcommand{\benchmemtask}[2]{%
  \stepcounter{benchtask}%
  \noindent\hangafter=1\hangindent=1.6em\textbf{\arabic{benchtask}.}~#1\par\vspace{0.15em}%
  \noindent\hspace{1.6em}\textit{Question:}~#2\\[0.55em]%
}

Our benchmark comprises 30 real-world mobile tasks grouped into five core capability categories plus a hybrid category that composes them, with five tasks each. Table~\ref{tab:benchmark-tasks} lists all tasks in capability order. Memory-management tasks (Category~5) are two-stage: stage~A executes the task, and stage~B asks memory-related questions.

\begin{longtable}{@{}>{\raggedright\arraybackslash}p{\textwidth}@{}}
\caption{Benchmark tasks grouped by capability category.}
\label{tab:benchmark-tasks} \\
\endfirsthead
\multicolumn{1}{c}{\tablename~\thetable\ — \textit{continued from previous page}} \\
\endhead
\multicolumn{1}{r}{\textit{Continued on next page}} \\
\midrule
\endfoot
\endlastfoot
\benchcat{Category 1 --- GUI Operation}
\benchtask{Open Tulsi Gallery, then open the first image whose picture contains the text AOHP, enter edit mode, and set Brightness to +57. Brightness setting completed, end the task without exiting the brightness setting interface.}
\benchtask{Open the Notes app. Change the app theme to Dark, then change Alignment to Center.}
\benchtask{Open Calendar and create an event on the 15th of this month titled Development Meeting with description AOHP Development Meeting, starting at 11:15 AM and ending at 1:30 PM.}
\benchtask{In Markor, merge \texttt{q1/notepad.md}, \texttt{q2/draft.md}, and \texttt{q3/minutes.md} in that order into \texttt{sprint\_summary.md} with a blank line between each file, then save.}
\benchtask{The Notes entry Team Roster contains Alice's mobile number. Create a new contact Alice with that number and text the number to HR at 10086.}
\benchcat{Category 2 --- Non-GUI Operation}
\benchtask{Delete exactly one Markdown file at \texttt{/sdcard/Documents/AOHP/benchmark/projects/archive/2024/q4/test.md}.}
\benchtask{Use Markor to open the Markdown file at \texttt{/sdcard/Documents/AOHP/benchmark/projects/archive/2024/q4/test.md}. Finish after the file is opened.}
\benchtask{Move \texttt{/sdcard/Documents/AOHP/benchmark/projects/archive/2024/month\_03/ledger.md} to \texttt{/sdcard/Documents/AOHP/benchmark/projects/archive/2024/month\_04/ledger.md}.}
\benchtask{Compute the product of all numeric cells in \texttt{/sdcard/Documents/AOHP/benchmark/data/metrics\_2024.csv} and write the integer result to \texttt{/sdcard/Download/metrics\_product.txt}.}
\benchtask{From \texttt{/sdcard/Documents/AOHP/benchmark/data/api\_metrics.json}, compute the P99 latency in milliseconds (rounded to integer) using the nearest-rank method on \texttt{samples\_ms}---sort ascending, let $n$ be the sample count, take the value at 0-based index $\lceil 0.99n \rceil - 1$---and write it to \texttt{/sdcard/Download/latency\_p99.txt}.}
\benchcat{Category 3 --- Event Capture}
\benchtask{HR will send an onboarding SMS shortly. When the notification arrives, read Alice's 11-digit mobile number from it and create a new contact Alice with that number.}
\benchtask{A bank SMS with a login verification code will arrive soon. After the notification appears, open Notes, create a note titled Bank OTP, and write only the 6-digit code in the body.}
\benchtask{In Clock, create a 5-minute timer and start it. Wait for the timer-finished notification, then send Alice an SMS saying ``Lunch has been heated and is ready''.}
\benchtask{Bob will text you soon. After Bob's SMS notification appears, open the camera and take a photo.}
\benchtask{Wait for the salary deposit bank notification. When it appears, send Alice an SMS saying ``Salary's in --- I'm treating you to a big dinner tonight!''}
\benchcat{Category 4 --- Multi-source Information Retrieval}
\benchtask{Please find and compile information for the next Development Meeting in the given apps---the start time (MM-DD HH:MM; the Calendar app does not show the year, so omit the year) of the Development Meeting event in the Calendar app, the first agenda item from Meeting Prep in the Notes app, and organizer Alice's mobile number in the Contacts app. Write three lines to \texttt{/sdcard/Documents/AOHP/answers/ir\_dev\_meeting\_brief.txt}.}
\benchtask{I'm preparing for a business trip to Paris. Please find and compile the following information in the given apps---the Paris Trip arrival date in Calendar, comma-separated items still to buy from Alice's SMS in Messages, and the total spent (integer USD) on \#Paris-tagged items in \texttt{travel/paris\_purchases.md} in the Markor app. Write to \texttt{/sdcard/Documents/AOHP/answers/ir\_paris\_trip\_pack.txt} (three lines---arrival date (MM-DD), shopping list, total amount as an integer).}
\benchtask{Please find and compile May reimbursement information in the given apps---the three amounts on \texttt{receipt\_aohp.png} in Gallery (\$12.50, \$8.00, \$15.75) and reimbursable entries for the same month in Markor \texttt{q3/minutes.md}. Write the total reimbursable amount (integer USD) to \texttt{/sdcard/Documents/AOHP/answers/ir\_reimbursement\_may\_total.txt}.}
\benchtask{Please find and compile the Board Review arrangement in the given apps---the meeting start time and duration from the Calendar app, and the meeting room from colleague Carl's latest SMS in the Messages app. Write one line to \texttt{/sdcard/Documents/AOHP/answers/ir\_board\_meeting\_confirm.txt} as \texttt{start\_time;room\_name} (time format MM-DD HH:MM).}
\benchtask{Please find and compile the shipment reminder in the given apps---the product name and order number for the pending shipment from the Mia Gift Order note in Notes, and Mia's phone number in Contacts. Write one line to \texttt{/sdcard/Documents/AOHP/answers/ir\_order\_mia\_shipment.txt} as \texttt{product\_name;order\_number;phone\_number} (phone\_number digits only).}
\benchcat{Category 5 --- Memory Management}
\benchmemtask{Open Calendar and create Development Meeting on the 15th of next month (11:15--13:30, description AOHP Development Meeting) and add Task AOHP Task under the 30th of next month (description Fork AOSP in aohp-os organization).}{What is the end time of the Development Meeting on the 15th of next month? What is the title of the Task on the 30th of next month? Answer in two lines.}
\benchmemtask{From the HR SMS, add contact Alice with mobile and extension.}{What are the last four digits of Alice's mobile number?}
\benchmemtask{In Markor, merge \texttt{/sdcard/Documents/AOHP/benchmark/projects/archive/2024/q1/notepad.md}, \texttt{q2/draft.md}, and \texttt{q3/minutes.md} in that order into \texttt{/sdcard/Documents/AOHP/benchmark/projects/archive/2024/sprint\_summary.md} with a blank line between each file, then save.}{What is the full filename of the third source file you merged?}
\benchmemtask{In Notes, open the note AOHP Backlog. Based on the todo items listed in the AOHP Backlog note, create a Checklist named AOHP TODO in the Notes app.}{What is the second todo item in the AOHP Backlog note?}
\benchmemtask{In Markor, open \texttt{/sdcard/Documents/AOHP/benchmark/projects/archive/2024/q4/test.md} by navigating from the Markor root.}{How many folder levels did you open from the Markor root to reach that file? Reply with a single number.}
\benchcat{Category 6 --- Hybrid (Capability Composition)}
\benchtask{Change the next Development Meeting in Calendar to 2 hours long, then set an alarm in Clock 15 minutes before that meeting's start time.}
\benchtask{Text Carl asking which room Board Review is in. While waiting, open Calendar and change the Board Review on the 20th of next month start time to 13:00. When his reply notification appears, set the event location to the room name from the reply.}
\benchtask{A colleague will SMS a meeting time. After the notification, check Calendar---if free, reply OK and schedule a 1-hour event; if busy, reply "Not available in this time slot." to the inviter and text the conflicting meeting organizer Sorry, I can't attend the meeting.}
\benchtask{In Notes checklist AOHP Cleanup, mark Archive Phoenix prototypes as done, then delete all Phoenix-related files from the Download folder.}
\benchtask{Family dinner this Saturday. Bob will send a menu SMS shortly; Notes Family Dinner lists drinks to prepare. Text Alice the wanted dishes and drinks in the format \texttt{dish1,dish2,...;drink1,drink2,...}}
\midrule
\end{longtable}

\end{document}